\documentclass[letterpaper]{article}
\usepackage{aaai19}
\usepackage{times}
\usepackage{helvet}
\usepackage{courier}
\usepackage{url}
\usepackage{graphicx}
\usepackage[titletoc,toc,title]{appendix}
\usepackage[utf8]{inputenc}
\usepackage{amsmath,amssymb,amsthm}
\usepackage{color}
\usepackage{algorithmic}
\usepackage[ruled,vlined,linesnumbered]{algorithm2e}
\usepackage{xspace}
\newcommand{\gmin}{\mbox{$g_{min}$}\xspace}

\newcommand{\fwmin}{\mbox{$f^W_{min}$}\xspace}
\newcommand{\fmin}{\mbox{$f_{min}$}\xspace}
\newcommand{\fw}[1]{\mbox{$f^W(#1)$}\xspace}
\newcommand{\f}[1]{\mbox{$f(#1)$}\xspace}

\newcommand{\OPEN}{\mathrm{Open}}

\newcommand{\astar}{{\sc A$^*$}\xspace}
\newcommand{\wastar}{{\sc wA$^*$}\xspace}

\newcommand{\fbound}{{\sc F}\xspace}

\newcommand{\lmcut}{{\sc LM-Cut}\xspace}
\newcommand{\ipdb}{{\sc iPDB}\xspace}
\newcommand{\cegar}{{\sc Cegar}\xspace}
\newcommand{\opcount}{{\sc Operator-Counting}\xspace}
\newcommand{\potential}{{\sc Potential}\xspace}
\newcommand{\saturated}{{\sc Cost-Partioning}\xspace}

\newtheorem{theorem}{Theorem}

\definecolor{brown}{cmyk}{0,0.5,1,0.2}
\definecolor{myGreen}{rgb}{0,0.8,0.4}
\definecolor{anothercolor}{rgb}{0,0.5,0}
\definecolor{donecolor}{rgb}{0.8,0.8,0.8}

\newcommand{\solved}{568\xspace}
\newcommand{\lastipc}{30\xspace}
\newcommand{\previousipc}{538\xspace}

\title{Error Analysis and Correction for Weighted A*'s Suboptimality (Extended Version)}

\author{Robert C. Holte\thanks{Work done while at Universidad Carlos
    III de Madrid}\\
  Computing Science Department\\
  University of Alberta \\
  Edmonton, Canada  T6G 2E8\\
  rholte@ualberta.ca\\
  \And Rubén Majadas, Alberto Pozanco \and Daniel Borrajo\\
  Departamento de Informática\\
  Universidad Carlos III de Madrid\\
  28911 Leganés (Madrid), Spain\\
  \{rmajadas,apozanco\}@pa.uc3m.es, dborrajo@ia.uc3m.es
  }
\def\year{2019}

\begin{document}

\maketitle

\begin{abstract}
  Weighted \astar (\wastar) is a widely used algorithm for rapidly, but suboptimally, solving planning and search
  problems.  The cost of the solution it produces is guaranteed to be at most $W$ times the optimal solution cost, where
  $W$ is the weight \wastar uses in prioritizing open nodes.  $W$ is therefore a suboptimality bound for the solution
  produced by \wastar.  There is broad consensus that this bound is not very accurate, that the actual suboptimality of
  \wastar's solution is often much less than $W$ times optimal.  However, there is very little published evidence
  supporting that view, and no existing explanation of why $W$ is a poor bound.  This paper fills in these gaps in the
  literature. We begin with a large-scale experiment demonstrating that, across a wide variety of domains and heuristics
  for those domains, $W$ is indeed very often far from the true suboptimality of \wastar's solution. We then
  analytically identify the potential sources of error. Finally, we present a practical method for correcting for two of
  these sources of error and experimentally show that the correction
  frequently eliminates much of the error.
\end{abstract}

\section{Introduction}\label{sec:intro}

In bounded suboptimal search, a bound $\beta \ge 1$ is given along with the problem to be solved, and the cost, $C$, of
the solution returned must be no more than $\beta C^*$, where $C^*$ is the problem's optimal solution cost.  In other
words, $\beta$ is an upper bound on the allowable suboptimality, $C/C^*$.

The most popular algorithm for bounded suboptimal search, and the focus of our paper, is Weighted \astar, \wastar
for short~\cite{Pohl1970aij}. $\beta$ is given to \wastar in the form of a weight $W$ that \wastar uses in its function
$f(s) = g(s) +Wh(s)$ for ordering nodes on the $\OPEN$ list.  The solutions returned by \wastar are guaranteed to cost
no more than $WC^*$ if the heuristic $h$ is admissible (p.~88~\cite{pearl84}, \cite{DavisBW88,Thayer2008}).

Although it is widely believed that $W$ is often a very loose upper bound on $C/C^*$ for \wastar's solutions, published
data supporting this belief is scarce and limited in its variety.  The first contribution of this paper
(Section~\ref{sec:CdivCstar}) is to provide compelling evidence supporting this belief
via a large-scale experiment involving six different
domain-independent heuristics (or combinations of them), and \solved problems drawn from 42 domains from the
International Planning Competition\footnote{http://ipc.icaps-conference.org} (IPC), and 400 problems from three non-IPC
domains.

We then identify four potential causes of $W$ being a loose upper bound (Section~\ref{sec:loose}).  For
two of these we present a practical method to correct for the error they introduce. This method is based on the actual
cost of the solution produced for the given problem and other information that is only available after the problem has
been solved, so it provides a ``post hoc'' suboptimality bound, in contrast to an {\it a priori} bound like $W$.  We
call this bound \fbound.

Finally (Section~\ref{sec:exp}), we repeat the large-scale experiment of Section~\ref{sec:CdivCstar} to examine how much
of $W$'s looseness has been eliminated by correcting for these errors.

\section{Preliminaries}\label{sec:prelim}

We assume the unweighted heuristic $h$ is admissible, but not necessarily consistent.
%Unless explicitly stated we do not assume $h$ is consistent.
We allow $h(s)=0$ for non-goal states and action costs of $0$. We assume $C^*>0$ and $W \ge 1$.

\f{s} denotes the unweighted $f$-value of state $s$, i.e. %\f{s}$=
$g(s)+h(s)$.  The weighted $f$-value, \fw{s}, is $g(s)+Wh(s)$.

\wastar is an iterative algorithm and certain key quantities can change from one iteration to the next.  For example,
\fwmin is the minimum weighted $f$-value of the nodes on
$\OPEN$ at the beginning of an iteration.  Other quantities defined over the nodes on
$\OPEN$ at the beginning of an iteration are:

\begin{itemize}
%% \item \fmin, the minimum unweighted $f$-value,
\item \gmin, the minimum $g$-value,
\item $g(s)$, the $g$-value of state $s$, and
\item $n_{opt}$ (defined below, Section~\ref{sec:loose}).
\end{itemize}

All these should have the iteration number as part of their notation but we found this cumbersome.  However,
when several of these quantities co-occur in a formula, they are referring to the respective quantities on the same
iteration.

\section{Experimental Setup}\label{sec:setup}

Our experiments aim to measure the accuracy of $W$ and the \fbound bound across a wide range of domains and heuristics.
We used $W \in \{1.2,1.5,2,4,8,16\}$. The experiments were run on Intel(R) Xeon(R) CPU X3470 @ 2.93GHz machines. We used
a time limit of 1800s and a memory limit of 8Gb.

Our primary experiments began with 732 of the 747 IPC problems\footnote{In the original SoCS'19 paper, we incorrectly
  reported 747. This number does not affect any of the remaining results of the paper.} solved optimally by the
lite-enhanced DM-HQ algorithm~\cite{MSlite2018} from 39 domains and the 110 solved problems from 6 domains in the 2018
IPC optimal track.\footnote{We selected the 7 domains without conditional effects, since some heuristics do not handle
  them. Then, we selected problems for which the upper and lower bounds were equal
  (https://bitbucket.org/ipc2018-classical/domains/src/default/), since we need the optimal solutions costs for
  computing $C/C^*$. {\tt petri-net-alignment} did not have these values updated, so we did not use it.}  Only \solved
of these problems (\previousipc from pre-2018 IPCs and \lastipc from the 2018 IPC) were solved within our time and
memory limits by all combinations of $W$-values and heuristics. The figures and discussion in this paper are based only
on these commonly solved problems.  We used Fast Downward's implementation of \wastar for experiments on these
domains~\cite{fast-downward-jair}.

The heuristics used for the IPC problems are high-quality admissible domain-independent heuristics from the planning
literature -- \lmcut~\cite{LMcut}, \ipdb~\cite{haslumBHBK07},\footnote{\ipdb was given 30 seconds to build its pattern
  database.}  \cegar~\cite{SeippH13}, \opcount~\cite{PommereningRHB14},\footnote{We used constraints \lmcut and state
  equations.}  \potential~\cite{SeippPH15},\footnote{Potentials for all facts, optimized for a high average heuristic
  value on all states~\cite{SeippPH15}.} and a method for combining heuristics, Saturated
\saturated~\cite{Seipp17}.\footnote{We have used their best option: diverse saturated cost partitioning over pattern
  database and Cartesian abstraction heuristics.}  \ipdb, \cegar, and \potential are guaranteed to be consistent, but
\lmcut and \opcount are not (the latter due to its internal use of \lmcut).

As secondary experiments, we used the 15-puzzle with the Manhattan Distance heuristic and the 100 standard test
instances~\cite{korf85}, the 15-pancake puzzle with the GAP heuristic~\cite{Helmert10} and two weakened versions of it,
GAP-1 and GAP-2~\cite{MMAAAI16}, and 200 randomly generated instances, and an industrial vehicle routing problem (VRP)
with the Minimum-Spanning-Tree heuristic and 100 randomly generated instances. For these we used our own implementations
of \wastar. All of these problems were solved by \wastar.

\section{How Far is $C/C^*$ from $W$ in Practice?}\label{sec:CdivCstar}

Almost no data has been published documenting how inaccurate $W$ is as an estimate of the suboptimality ($C/C^*)$ of
\wastar's solutions.  One of the rare exceptions is Table~2 in~\cite{Korf1993}, which shows the average solution cost
produced on 100 15-puzzle instances as $W$ varies from $3$ to $99$.  Concerning \wastar and the other algorithms he is
studying, Korf observes ``for small values of $W$, they produce nearly optimal solutions whose lengths grow very slowly
with $W$.''

\begin{figure}[t!]
%%   \hspace*{-0.4cm}
  \begin{tabular}{cc}
    \potential &
                 \begin{minipage}{0.6\textwidth}
                   \includegraphics[scale=0.32]{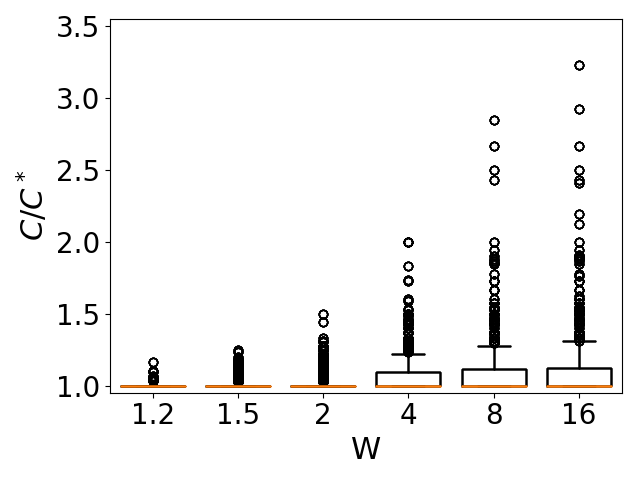}
                 \end{minipage}\\
   VRP &
           \begin{minipage}{0.6\textwidth}
             \includegraphics[scale=0.32]{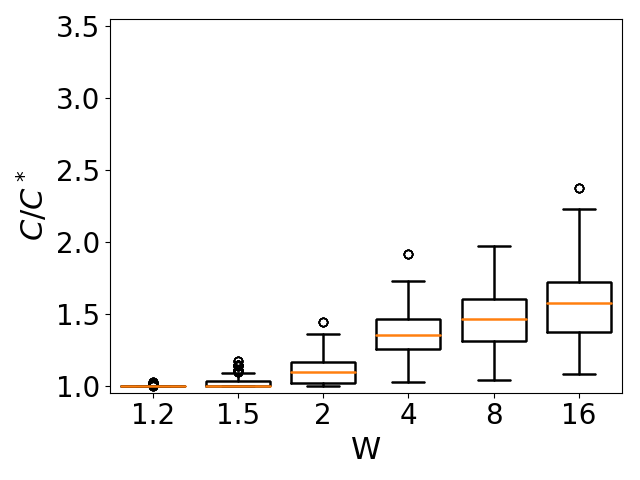}
           \end{minipage}
  \end{tabular}

    \caption{$C/C^*$ as a function of $W$ using the \potential heuristic (IPC domains) and for the VRP domain.}
  \label{fig:CoverCstar}
\end{figure}

Figure~\ref{fig:CoverCstar} shows that Korf's observations hold across a broad range of domains and heuristics.
Boxplots\footnote{The lower and upper edges of the box represent the first (Q$_1$) and third (Q$_3$) quartiles of the
  data, respectively. The whisker extends above the box to the highest data point below Q$_3 + 1.5($Q$_3-$Q$_1)$.  Data
  points beyond the whisker are shown individually. The solid horizontal line (orange) inside the box is the median.}
show the distribution of $C/C^*$ values for our \solved IPC problems for all values of $W$. This figure is for a
specific heuristic -- \potential (IPC domains) -- and a specific non-IPC domain (VRP). The other $C/C^*$ plots are very
similar except as noted below.  The key features for IPC domains are:
%%  of Figure~\ref{fig:CoverCstar}

\begin{itemize}
	
\item In all cases at least $25\%$ of the problems are solved optimally.

\item For all heuristics except \cegar, the median value of $C/C^*$ is 1.0 for all values of $W$.  For \cegar, the
  median is 1.0 for $W=1.2$ and rises very slowly as $W$ increases to a maximum of 1.1 for $W=16$.

\item For all values of $W$, $C/C^*$ is less than 1.15 on $75\%$ of the problems for all heuristics except for \cegar,
  for which the $75^{th}$ percentile is 1.3.

\item $C/C^*$ is almost always less than $\sqrt{W}$.  In particular: 73 of the \solved\ $C/C^*$ values are greater than
  $\sqrt{1.2}$ when $W=1.2$, 47 are greater than $\sqrt{1.5}$ when $W=1.5$, 25 are greater than $\sqrt{2}$ when $W=2$, 8
  are greater than $\sqrt{4}$ when $W=4$, 3 are greater than $\sqrt{8}$ when $W=8$, and none is greater than $\sqrt{16}$
  when $W=16$.
\end{itemize}

\noindent
The results in the non-IPC domains are a bit different than in the IPC domains.
%% Figure~\ref{fig:CoverCstar} also shows the distribution of values of $C/C^*$ as a function of $W$ in the VRP domain.
The results shown in Figure~\ref{fig:CoverCstar} for VRP are representative of the distributions for the other non-IPC
domains, with the 15-puzzle being somewhat worse, and the Pancake puzzle being somewhat better for all of its
heuristics.  In contrast to the IPC domains, the median, and even the lower quartile (bottom of the boxes), is greater
than 1.0 for $W \geq 2$. On the other hand, the largest values of $C/C^*$ are smaller in the non-IPC domains than in the
IPC domains for all values of $W$.
  
The observations in these experiments are consistent with Korf's and others'~\cite{Thayer2008}, and
provide compelling evidence that $W$ is a poor estimate of the suboptimality of \wastar's solutions, especially when $W$ is large.

\section{Why is $W$ a Loose Bound?}\label{sec:loose}

Although it is possible to construct examples in which $C/C^* =W$, we have just seen that this virtually never happens
in practice.  In this section, we identify potential causes of this
by examining a typical proof that $C/C^* \le W$.  This proof uses the fact that, at the beginning of any iteration, for
any optimal path $P$ from $start$ to any unclosed state -- in particular, any goal state -- there exists a node,
$n_{opt} \in P$, on $\OPEN$ such that $g(n_{opt}) = g^*(n_{opt})$ (Lemma~1~\cite{astar68}).\footnote{Hart et al. proved
  this lemma in the context of A* but it applies much more broadly, including to \wastar.} Then

\begin{center}
\begin{tabbing}
$\fwmin$ \= $\le \fw{n_{opt}} = g(n_{opt}) + Wh(n_{opt})$  \hspace*{0.7cm} \= (1) \\
%\>        $= g^*(n_{opt})  + Wh(n_{opt}) \le g^*(n_{opt})  + Wh^*(n_{opt})$\\
\>        $= g^*(n_{opt})  + Wh(n_{opt})$ \> (2)\\
\>        $\le g^*(n_{opt})  + Wh^*(n_{opt})$ \> (3) \\
%\>        $\le Wg^*(n_{opt})  + Wh^*(n_{opt}) \hspace*{1.5cm}(*)$\\
\>        $\le Wg^*(n_{opt})  + Wh^*(n_{opt})$ \> (4)\\
\>        $= W(g^*(n_{opt})  + h^*(n_{opt}))$  \> (5) \\
\>        $= WC^*$.
\end{tabbing}
\end{center}

\noindent
On \wastar's final iteration, \fwmin$=C$, so this derivation establishes that $C/C^* \le W$.

One reason $W$ can overestimate the true value of $C/C^*$ is that this derivation applies to \fwmin on every iteration,
it is not specific to the last iteration.\footnote{Theorem~2 by Thayer and Ruml~\shortcite{Thayer2008} also makes this
  observation but they do not exploit it in any way.}  A tighter bound on $C/C^*$ will almost always be obtained by
considering the largest value of \fwmin that occurred throughout \wastar's execution.  We call this value $F$.  It is
easily computed during search and, as we shall see, it can be used to define a much better bound for $C/C^*$.

The other sources of potential error (overestimation) in this derivation are the steps that involve an inequality:
lines~(1), (3), and (4).  We do not see any practical way of correcting for the error introduced by line~(1), since
nodes on an optimal solution path cannot be identified during or even after \wastar's search.  Furthermore, a step
introducing $n_{opt}$ seems inevitable in any derivation of a relation between $C$ and $C^*$, since it is via $n_{opt}$
that $C^*$ eventually emerges in the derivation.

The error introduced in line~(3) is caused by the inaccuracy of the heuristic function $h$. Again, we see no practical
way of correcting this error based on the information available during or after \wastar's search. And similar to the
introduction of $n_{opt}$ in line~(1) we see no way of avoiding introducing $h^*(n_{opt})$ in deriving a relation
between $C$ and $C^*$.

The situation is different with the error introduced in line~(4).  Multiplying $g^*(n_{opt})$ by $W$ has no intrinsic justification;
it is only done to allow the equation to be simplified.  A better way to proceed is as follows (the first few
steps are the same as before and are not shown):

\begin{align*}
%f_{min}^t & \le g(n^t_{opt}) + Wh(n^t_{opt})\\
%& = g^*(n^t_{opt})  + Wh(n^t_{opt})\\ 
\fwmin & \le g^*(n_{opt})  + Wh^*(n_{opt})\\
& = g^*(n_{opt})  + W(C^*-g^*(n_{opt}))\\ 
& = WC^* - (W-1)g^*(n_{opt})\\
& = WC^* - (W-1)g(n_{opt})\\
& \le WC^* - (W-1)\gmin.
\end{align*}

This derivation replaces the potentially very large error introduced in the first derivation by multiplying
$g^*(n_{opt})$ by $W$ with the error seen in the final line: replacing $g(n_{opt})$ by \gmin. In preliminary experiments
this error was 0 more than 50\% of the time and was rarely a significant fraction of the total error.

As noted above, this derivation is true for \fwmin on all iterations, and therefore we have
\[ F \le WC^* - (W-1)\gmin \]

\noindent
where $\gmin$ is the smallest $g$-value on $\OPEN$ at the beginning of an iteration when a node $n$ with $f^W(n) = F$ was removed from $\OPEN$.  It is interesting to
see an additive correction term for $F \le WC^*$ that will only be $0$, when $W > 1$, in the rare situation that $\gmin=0$.  With some
simple algebraic rearrangement, this inequality gives the following suboptimality bound for \wastar, which we call the
\fbound bound:

\begin{equation*}
\label{fbound}
\frac{C}{C^*} \le \frac{CW}{F + (W-1)\gmin}.
\end{equation*}

\begin{figure}[b!]
  \centering
  \includegraphics[scale=0.3]{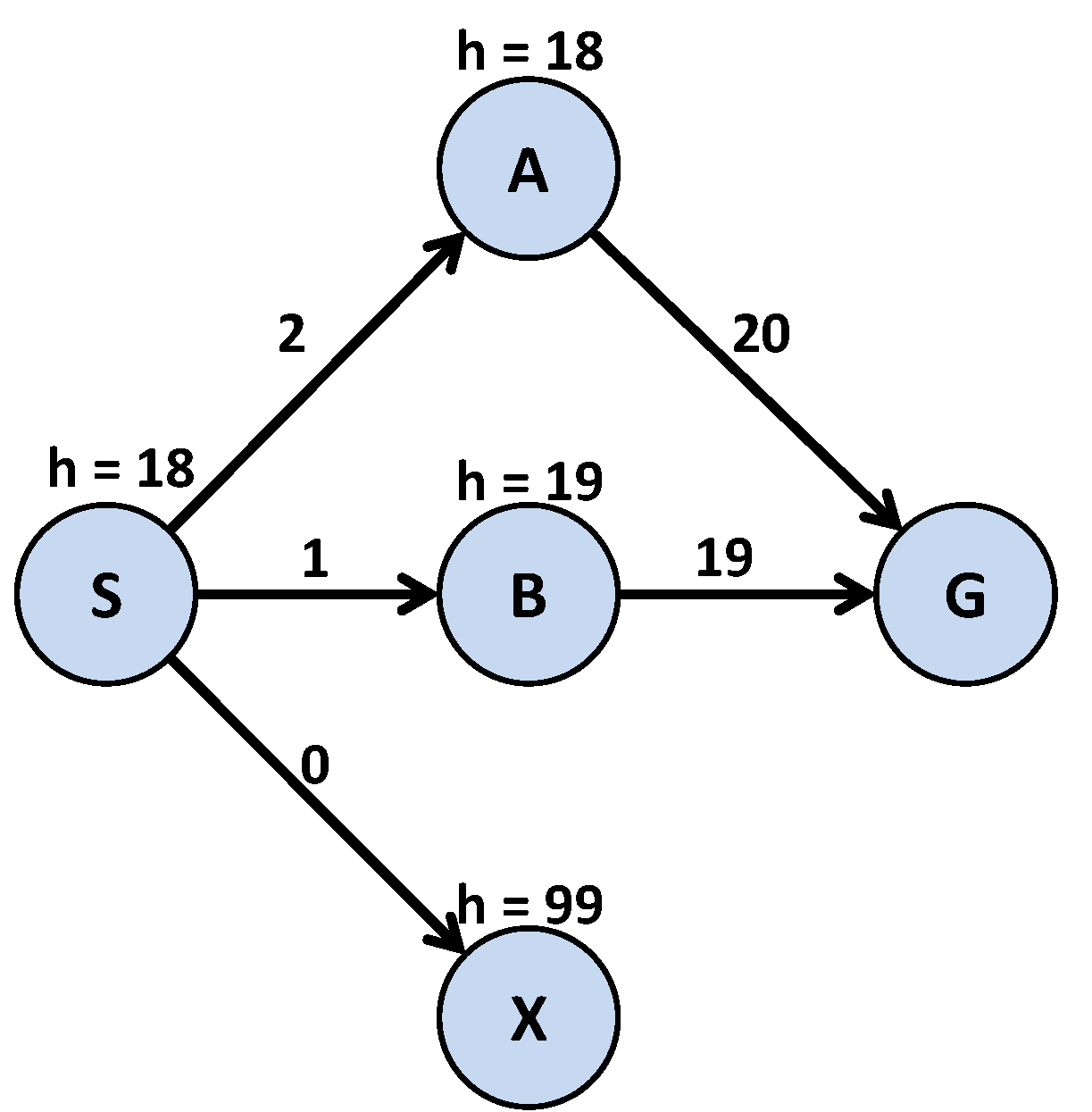}
  \vspace*{-0.2cm}
  \caption{The \fbound bound is a much better bound than $W$.}
  \label{fig:example5}
\end{figure}
 
To see how accurate the \fbound bound can be, consider Figure~\ref{fig:example5} when $W=10$.
The heuristic values here are consistent and \gmin is $0$ on all iterations.
The optimal path is through $B$, and its cost is $C^*=20$. $B$'s large $h$-value prevents \wastar from expanding $B$ and
finding the optimal path.  Instead, \wastar returns the path through $A$, costing $22$.
$F = \max\{f(S),f(A),f(G)\} = 182$, so the \fbound bound is $(22 \cdot 10)/182 = 1.2$, which is very close to the actual
value of $C/C^*$ ($22/20 = 1.1$) and almost an order of magnitude smaller than $W$.
We have experimentally observed that computing the \fbound bound results in a negligible overhead.

\section{Experimental Evaluation}\label{sec:exp}

The \fbound bound corrects for two sources of overestimation in using $W$ as a bound on $C/C^*$. The question addressed
by the experiments in this section is, how effective are those corrections? Do they eliminate much of the
overestimation?  We will answer this question by evaluating the accuracy of the \fbound bound relative to the accuracy
of $W$ as an upper bound on $C/C^*$.

To quantify the accuracy of the \fbound bound we must take into account how its value, $v_{\fbound}$, on a given problem
compares to the minimum and maximum possible values on the problem ($C/C^*$ and $W$ respectively).  We denote the
accuracy of the \fbound bound's value by $\rho$, defined as\footnote{If $W = C/C^*$ then we define $\rho$ to be $0$.}
%\footnote{This assumes $W > C/C^*$. If  $W = C/C^*$ we define $\rho = 0$ in this case.}
\[ \rho = \frac{\log(v_{\fbound})-\log(C/C^*)}{\log(W)-\log(C/C^*)}.    \]
\noindent
The denominator is the distance, in log space, between the \fbound bound's minimum and maximum possible values, and the
numerator is the distance in log space between the \fbound bound's actual value on a problem ($v_{\fbound}$) and its
minimum possible value. $\rho$ is always between $0$ and $1$ and represents the \fbound bound's distance from $C/C^*$ as
a fraction of $W$'s distance from $C/C^*$. A smaller value means better accuracy.

We consider $\rho> 0.5$ to be a poor score since it means $v_{\fbound}$ is closer to $W$, in log space, than it is to
$C/C^*$. We consider $\rho \le 0.25$ to be a good score.  For example, if $W=32$ and $C/C^* = 2$, $v_{\fbound}=8$
corresponds to $\rho = 0.5$ while $v_{\fbound}=4$ corresponds to $\rho = 0.25$.
 
Figure~\ref{fig:ROBplot} has a boxplot for $\rho$ ($y$-axis) as a
function of $W$ ($x$-axis) for the \ipdb and \potential heuristics.
The results for other heuristics are similar to the \ipdb ones, except for the \saturated heuristic where they are
better.  The \potential heuristic results are uniformly poorer than the rest.
The main trends for all heuristics except for \potential are:
%% in Figure~\ref{fig:ROBplot}

\begin{figure}[htb]
%%   \centering \ipdb \vspace*{-0.2cm} \includegraphics[scale=0.4]{fig3/F-rho-ipdb}
%% 
%%   \potential \vspace*{-0.2cm} \includegraphics[scale=0.4]{fig3/F-rho-all_states}

%%   \hspace*{-0.4cm}
  \begin{tabular}{cc}
    \ipdb &
            \begin{minipage}{0.6\textwidth}
              \includegraphics[scale=0.3]{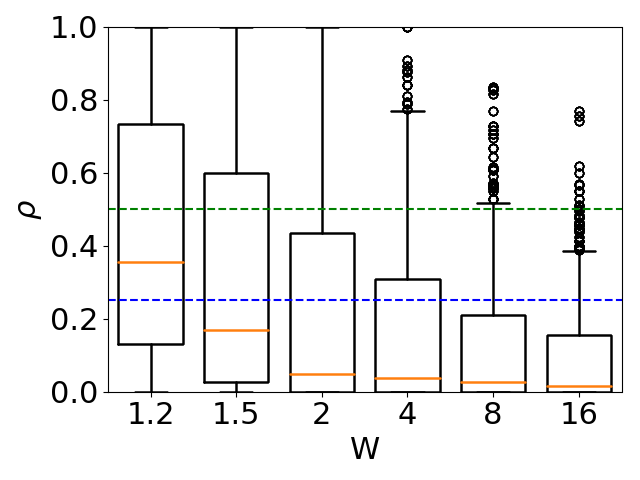}
            \end{minipage}\\
    \potential &
            \begin{minipage}{0.6\textwidth}
              \includegraphics[scale=0.3]{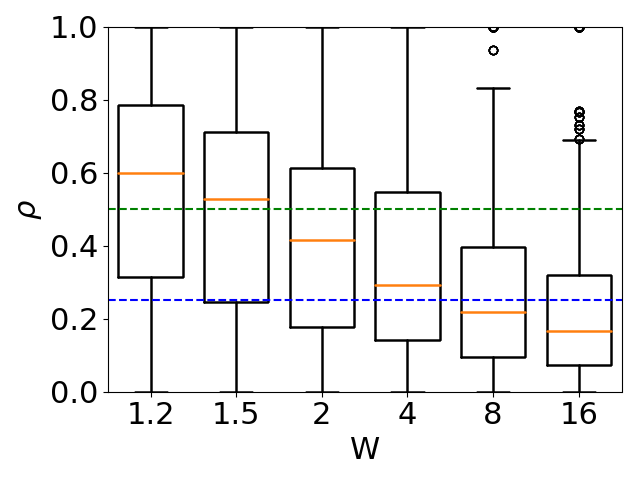}
            \end{minipage}
%%     \change{(a) \ipdb} & \change{(b) \potential}
  \end{tabular}

  \vspace*{-0.2cm}
%%   \centering
%%   \includegraphics[scale=0.37]{fig3/F-rho-ipdb}
%%   \vspace*{-0.2cm}
  \caption{Boxplots showing the accuracy ($\rho$) on the $y$-axis of
    the \fbound bound as a function of $W$ ($x$-axis) for the \ipdb
    and \potential heuristics.  The horizontal lines that span the
    entire $x$-axis indicate our thresholds for ``good''
    ($\rho \le 0.25$) and ``bad'' ($\rho\ge 0.5$) $\rho$ values.}
%%   \caption{Boxplots showing the accuracy ($\rho$) on the $y$-axis of the \fbound bound as a function of $W$ ($x$-axis)
%%     for the \ipdb heuristic.  The horizontal lines that span the entire $x$-axis indicate our thresholds for ``good''
%%     ($\rho \le 0.25$) and ``bad'' ($\rho\ge 0.5$) $\rho$ values.}
  \label{fig:ROBplot}
\end{figure}

\begin{itemize}
	
\item[(1)] The accuracy of the \fbound bound improves as $W$ increases.  All aspects of the distribution improve: the
  median, the $75^{th}$ percentile, (top of the box), the upper whisker, and even the outliers. These all decrease as
  $W$ increases. This is also true in the non-IPC domains, except for VRP where the distributions only worsen
    from $W=1.2$ to $W=1.5$.
  
\item[(2)] In Figure~\ref{fig:ROBplot} the boxes are seated on $\rho=0$ for $W \geq 2$, meaning the \fbound bound
  perfectly predicts $C/C^*$ for at least $25\%$ of the problems when $W \ge 2$. This holds for all heuristics except
  \potential. In the non-IPC domains, the \fbound bound does not provide any perfect prediction except for a single
  Pancake puzzle problem with the GAP heuristic.

\item[(3)] For all heuristics except the \potential heuristic, the median values of the \fbound bound are ``good'' for
  all values of $W\geq 2$. When using \saturated, in 50\% of the problems the prediction is perfect when $W\geq 2$.
In the non-IPC 15-puzzle and VRP domains, the median values of the \fbound bound are ``good'' when $W \geq 8$. For the
Pancake puzzle the median value is ``good'' when $W \geq 4$ if the GAP heuristic is used, when $W \geq 8$ if the GAP-1
heuristic is used, and only when $W \geq 16$ if the GAP-2 heuristic is used.

\item[(4)] The distributions are very broad, covering the entire range of possible $\rho$ values for $W \le 4$. With the
  \opcount, \ipdb, and \saturated heuristics the $75^{th}$ percentile (top of each box) is a ``good'' $\rho$ value for
  larger values of $W$. This is also true in the non-IPC domains for $W=16$.  The tails of the distributions stretch
  into the region of ``poor'' $\rho$ values for all heuristics and $W$ values
  in the IPC domains but only when $W\leq 4$ for the non-IPC domains.

\end{itemize}

\section{Conclusions}

In this paper we have presented compelling evidence that $W$ is, indeed, almost always a very loose bound on the
suboptimality ($C/C^*$) of \wastar's solutions, especially for larger $W$.  We have also identified the causes of this
looseness and presented a practical method, the \fbound bound, for correcting for two of them once \wastar has finished
executing. Finally, we examined how effective these corrections are, i.e. how much more accurately $C/C^*$ is predicted
by the \fbound bound than by $W$.  Our overall conclusion is that the corrections embodied in the \fbound bound are very
effective (the \fbound bound predicts $C/C^*$ much more accurately than $W$) for any of the heuristics we tested when
$W \ge 8$.  However, there do exist problems for which the \fbound bound's predictions are poor, with the number of such
problems decreasing as $W$ increases.

We do not claim that the \fbound bound is the best possible post hoc suboptimality bound for \wastar. Indeed, we know it
is not when the heuristic being used is consistent, because we have proven, in that case, that the \fbound bound is
dominated by a bound based on the largest unweighted $f$-value on \wastar's open list upon termination (see Appendix~\ref{app:f1domF}).
Our aim in introducing the
\fbound bound is to show that it is possible to explain, and directly correct for, a great deal of $W$'s looseness as a
suboptimality bound for \wastar.

% !TeX spellcheck = en_CA

%\titleformat{\section}{\large\bfseries}{\appendixname~\thesection .}{0.5em}{}
\begin{appendices}
\section{The $\boldsymbol{f}$ bound dominates the $\boldsymbol{F}$ bound when the heuristic is consistent}\label{app:f1domF}

In this appendix \fmin refers to the smallest \emph{unweighted} $f$-value on $Open$ at the beginning of \wastar's last iteration, and the $f$ bound is defined to be $C/\fmin$.

\begin{theorem}
	The $F$ bound is dominated by the $f$ bound if the heuristic is consistent.
\end{theorem}

\noindent
{\it Proof:}
We have to prove that
	\begin{equation}\label{eq:f1dominatesF}
	\frac{CW}{F+(W-1)\gmin} \ge C/\fmin,
	\end{equation}
\noindent
always holds when $h$ is consistent.

Equation~\ref{eq:f1dominatesF} is equivalent to
\begin{equation}\label{eq:f1dominatesF2}
 W\fmin \ge F+(W-1)\gmin
\end{equation}

We observe that $F$ is always equal to the maximum $f^W$-value on the solution returned by \wastar; i.e. if the path returned is $S=s_0,s_1,\dots,s_n~(s_0=start, s_n$ a goal state$)$ and
$F_S = \max\{\fw{s_i}\}$, then $F=F_S$.  The reason is the following. $F < F_S$ is impossible
since all the nodes on $S$ are removed from $\OPEN$ before termination.  $F > F_S$ is impossible because at the start of
any iteration there is an $s_i$ on $\OPEN$ so $\fwmin \le F_S$ always holds. Therefore, $F=F_S$.

Let $m$ be a state on path $S$ with $\fw{m} = F$, let $n$ be a node on $Open$ at the beginning of \wastar's last iteration with $\f{n}=\fmin$, and let $P$	be the path by which $n$ was placed on $Open$ for the final time (i.e. with $\f{n}=\fmin$).

Note that, for all states $a$, $W\f{a} = (W-1)g(a) + \fw{a}$, so $W\f{a} \ge \fw{a}$ if $W>1$.

There are two cases to consider: (1)~$m$ is on path $P$, i.e. $n$ is a descendant of $m$; (2)~$m$ is not on path $P$.
	
\smallskip
{\bf Case 1. $\boldsymbol{m}$ is on path $\boldsymbol{P}$.} In this case, Equation~\ref{eq:f1dominatesF2} follows directly from the heuristic's consistency, as follows.	
\begin{align*}
h(m) & \le c(m,n) + h(n)   \\
	g(m) + h(m) & \le \f{n} = \fmin \\ 
	Wg(m) + Wh(m) & \le W\fmin \\  
	(W-1)g(m) + g(m)+Wh(m) & \le W\fmin \\  
	(W-1)g(m) + \fw{m} & \le W\fmin \\ 
	(W-1)g(m) + F & \le W\fmin \\ 
	(W-1)\gmin + F & \le W\fmin.
\end{align*}
	
\noindent
The last step of the derivation follows from the definition of $g_{min}$ (at the time $m$ was chosen for expansion).
	
\smallskip
\noindent {\bf Case 2. $\boldsymbol{m}$ is not on path $\boldsymbol{P}$.} This case has two subcases: (a)~$n$ was on $\OPEN$ with $\f{n}=\fmin$ at the time $m$ was chosen for expansion; (b)~$n$ was not on $\OPEN$ with $\f{n}=\fmin$ at the time $m$ was chosen for expansion.

\smallskip
\noindent {\bf Case~2(a). $\boldsymbol{n}$ was on $\boldsymbol{\OPEN}$ with $\boldsymbol{\f{n}=\fmin}$ when $\boldsymbol{m}$ was removed from $\boldsymbol{\OPEN}$.} Since $m$ was chosen for expansion, it must be that $\fw{n} \ge \fw{m} = F$. Noting that $W\fmin = W\f{n} = (W-1)g(n) + \fw{n} \ge (W-1)g(n) + F$ and that $g(n) \ge \gmin$ we get the desired result	($W\fmin \ge F+(W-1)\gmin$).
	
\smallskip
\noindent {\bf Case~2(b). $\boldsymbol{n}$ was added to $\boldsymbol{\OPEN}$ with $\boldsymbol{\f{n}=\fmin}$ after $\boldsymbol{m}$ was removed from $\boldsymbol{\OPEN}$.}  This means some
	state $x \ne n$ on path $P$ was on $\OPEN$ at the time $m$ was chosen for expansion (note: $x=m$ is not possible, since in the case we are considering $m$ is not on path $P$).
$\fw{x} > \fw{m}=F$ is impossible, since $x$ is eventually expanded (in	order to complete path $P$) and, by definition of $F$, no state $s$ is ever expanded with $\fw{s} > F$. $\fw{x} < \fw{m}$	contradicts $m$ being chosen for expansion, so at the time $m$ is removed from $\OPEN$, we must have 	$\fw{x} = \fw{m} =F$.  With this equality established, we can now use exactly the same reasoning as we did in Case~1, just substituting $x$ for $m$ everywhere:
\begin{align*}
h(x) & \le c(x,n) + h(n)   \\
g(x) + h(x) & \le \f{n} = \fmin \\ 
%Wg(x) + Wh(x) & \le W\fmin \\  
\textrm{... etc. ... } \\  
(W-1)\gmin + F & \le W\fmin.  ~~~~~~~~~~~~~~~~~~~~~~~~~~~~\qed
\end{align*}

%\section{Comparison of the Bounds}\label{sec:analysis}

\begin{figure}[h!]
	\centering
	\includegraphics[trim=4.0cm 6.1cm 5.9cm 2.3cm, clip,width=5.8cm]{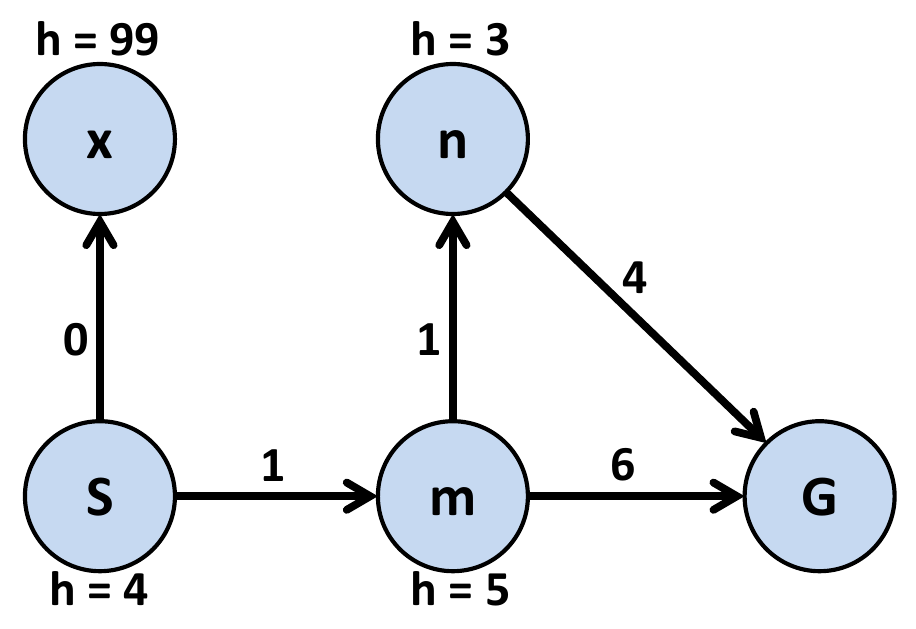}
	\caption{Inconsistent heuristic ($h(m) \nleq c(m,n)+h(n)$).}
%\caption{Example where the $F^-$ bound is better than the $f^1$ bound.}
	\label{fig:example8}
\end{figure}

Figure~\ref{fig:example8} shows that
the $f$ bound can be worse than the $F$ bound if the heuristic is even ``slightly" inconsistent.
$S$ is the start state, $G$ the goal. Consistency requires $h(m)-h(n) \le 1$, but it is 2.  $g_{min}$ is $0$ on all iterations.
The optimal solution, $S$--$m$--$n$--$G$, costs $C^* = 6$. When $W=2$, $\fw{n}=8$
%is large enough to prevents $n$ from being expanded.
so \wastar returns
% the suboptimal 
solution $S$--$m$--$G$ costing $C=7$.  $\fmin=\f{n}=5$ and $F = \fw{m}=11$.
%and $\gmin$, at the time $m$ was expanded, was $g(m)=1$.  The $F$ bound, $\frac{CW}{F+(W-1)\gmin} = \frac{14}{11+1} = \frac{7}{6}$, is exactly equal to $C/C^*$ and strictly smaller than $C/f^1 = \frac{7}{5}$.
%In fact, even
The $F$ bound ($\frac{CW}{F} = \frac{14}{11}$) is strictly smaller than the $f$ bound ($\frac{7}{5}$).
%In our experiments we see this occur on a few problems when the lmcut and operator counting heuristics are used.

\end{appendices}

\section*{Acknowledgements}

We would like to thank reviewers for their helpful comments. This work was partially funded by an Chair of Excellence
UC3M-Santander and by grants TIN2017-88476-C2-2-R and RTC-2016-5407-4 funded by Spanish Ministerio de Economía,
Industria y Competitividad.

\bibliographystyle{aaai}
\bibliography{Neil}

\end{document}